%% file: main.tex
\documentclass[runningheads]{llncs}
\usepackage[T1]{fontenc}
\usepackage{graphicx}
\usepackage{booktabs}
\usepackage[misc]{ifsym}
\newcommand{\corr}{(\Letter)}
\input{misc/packages}
\def\BibTeX{{\rm B\kern-.05em{\sc i\kern-.025em b}\kern-.08em
    T\kern-.1667em\lower.7ex\hbox{E}\kern-.125emX}}

\newcommand{\dddANNLearnedLayerDist}{3D MLP\xspace}
\newcommand{\ANNBaselineX}{MLP Baseline\xspace}

\newcommand{\dddSNNLearnedLayerDist}{3D SNN\xspace}
\newcommand{\SNNBaselineX}{SNN Baseline\xspace}

\newcommand{\dddANNfreeZ}{Relaxed 3D MLP\xspace}

\begin{document}

% \title{Training Neural Networks Through Topology Optimization
% }
\title{Training Neural Networks by \\Optimizing Neuron Positions
}

% \titlerunning{Training Neural Networks Through Topology Optimization}
\titlerunning{Training Neural Networks by Optimizing Neuron Positions}

\author{
Laura Erb\inst{1,2}%\thanks{These authors contributed equally and share first authorship.}
, Tommaso Boccato\inst{3}%$^\star$
\corr
, Alexandru Vasilache\inst{1,2}%$^\star$
, 
\\
Juergen Becker\inst{2}%\thanks{These authors share senior authorship.}
, Nicola Toschi\inst{3,4}%$^{\star\star}$
\\
\email{laura.erb1@gmx.de, tommaso.boccato@uniroma2.it, vasilache@fzi.de, juergen.becker@kit.edu, nicola.toschi@uniroma2.eu}
}
%\author{Anonymous Submission}

\authorrunning{Laura Erb et al.}
%\authorrunning{Anonymous Submission}

\institute{FZI Research Center for Information Technology, Karlsruhe, Germany 
\and
Karlsruhe Institute of Technology, Karlsruhe, Germany 
\and
Department of Biomedicine and Prevention, University of Rome Tor Vergata, Rome, Italy
\and
A.A. Martinos Center for Biomedical Imaging and Harvard Medical School, Boston, USA
}
%\institute{Anonymous Submission}

\maketitle

\input{misc/acronyms}
\begin{abstract}
\input{sections/abstract}

% \keywords{Topo\-logy Optimization, Neural Architecture Optimization, Spatial Embedding, Spiking Neural Networks}
\end{abstract}

\section{Introduction}
\input{sections/introduction}

\section{Methods}
\input{sections/methods}

\section{Experiments and Results}
\input{sections/results}

\section{Discussion}

\input{sections/discussion}

\section{Conclusion}
\input{sections/conclusion}

\subsubsection{Disclosure of Interests}
The authors have no competing interests to declare that are relevant to the content of this article.

\subsubsection{Acknowledgment}
The idea for this work arose from a collaboration at the  Capocaccia Cognitive Neuromorphic Engineering Workshop 2024.

\bibliographystyle{splncs04}
\bibliography{misc/literature}

% \newpage
% \appendix
% \section*{Appendices}
% \input{sections/sup}

\end{document}

%% file: misc/packages.tex
\usepackage{cite}
\usepackage{amsmath,amssymb,amsfonts}
\usepackage{algorithmic}
\usepackage{graphicx}
\usepackage{textcomp}
\usepackage{xcolor}
\usepackage{hyperref}

\usepackage[nolist]{acronym}
\newcommand{\snn}[1]{\hyperref[snn:#1]{P#1}}
\usepackage{tikz}
\usetikzlibrary{trees,positioning}
\usepackage{tabularx}
\usepackage{textcmds}
\usepackage{subcaption}
\usepackage{xspace}
\usepackage[utf8]{inputenc}
\usepackage[T1]{fontenc}     % Recommended for proper font encoding
\usepackage{cleveref}

%% file: misc/acronyms.tex
\begin{acronym}[Longest Abrev] %see https://www.namsu.de/Extra/pakete/Acronym.html
\acro{ANN}{Artificial Neuronal Network}
\acro{DBN}{Deep Belief Network}
\acro{CNN}{Convolutional Neural Network}
\acro{DNN}{Deep Neural Network}
\acro{LIF}[LIF]{Leaky Integrate-and-Fire}
\acro{IF}[IF]{Integrate-and-Fire}
\acro{LI}[LI]{Leaky Integrate}
\acro{ALIF}{Adaptive Leaky Integrate-and-Fire}
\acro{adex}[AdEx]{adaptive exponential integrate-and-fire}
\acro{LSTM}{Long Short-Term Memory}
\acro{RNN}{Recurrent Neural Network}
\acro{LSNN}{Long Short-Term Spiking Neural Network}
\acro{NN}{Neural Network}
\acro{SNN}{Spiking Neural Network}
\acro{ML}{Machine Learning}
\acro{DL}{Deep Learning}
\acro{AI}{artificial intelligence}
\acro{BPTT}{Backpropagation Through Time}
\acro{BP}{Backpropagation}
\acro{MLP}{Multi-layer Perceptron}

\end{acronym}

%% file: sections/abstract.tex
The high computational complexity and increasing pa\-ra\-me\-ter counts of deep neural networks pose significant challenges for deployment in resource-constrained environments, such as edge devices or real-time systems. To address this, we propose a parameter-efficient neural architecture where neurons are embedded in Euclidean space. During training, their positions are optimized and synaptic weights are determined as the inverse of the spatial distance between connected neurons. These distance-dependent wiring rules replace traditional learnable weight matrices and significantly reduce the number of parameters while introducing a biologically inspired inductive bias: connection strength decreases with spatial distance, reflecting the brain's embedding in three-dimensional space where connections tend to minimize wiring length. We validate this approach for both multi-layer perceptrons and spiking neural networks. Through a series of experiments, we demonstrate that these spatially embedded neural networks achieve a performance competitive with conventional architectures on the MNIST dataset. Additionally, the models maintain performance even at pruning rates exceeding 80\% sparsity, outperforming traditional networks with the same number of parameters under similar conditions. Finally, the spatial embedding framework offers an intuitive visualization of the network structure.

%% file: sections/introduction.tex
In scenarios requiring real-time inference, enhanced user privacy, and reduced reliance on cloud-based systems, it is often preferred to deploy \acp{DNN} directly on edge devices. However, the increasing computational complexity and parameter counts of modern \acp{DNN} present a significant challenge for deployment in resource-constrained environments \cite{Mobile_Device_Compression}. Addressing this challenge requires energy-efficient neural architectures with reduced memory footprints. 
\acp{SNN} offer a promising solution to meet low-energy consumption requirements. Their biologically inspired, spike-based computations enable highly efficient processing while maintaining high accuracy \cite{Embodied_neuromorphic_intelligence}. Nonetheless, in addition to energy efficiency, memory constraints on edge devices require further optimization techniques to reduce the number of parameters.

Incorporating biologically inspired inductive biases into the network architecture can reduce network complexity and parameters while guiding learning and improving generalization \cite{Inductive_Biases,BOCCATO2024127058,BOCCATO2024215}. For example, \acp{CNN} effectively reduce parameters by mimicking the concept of local receptive fields in the visual cortex \cite{MNIST}. Recent research showed that optimizing wiring rules based on gene expressions, rather than training individual weights, can improve learning efficiency and generalization while maintaining performance \cite{Prior_gene_expression}. This approach was inspired by %how genes efficiently encode connectivity patterns. 
the fact that certain neural circuits in the brain are prewired, which means their connectivity is genetically encoded rather than learned through experience \cite{Func_from_innate_wiring}. However, the complexity of these circuits far exceeds the information capacity of the genome, a challenge known as the genomic bottleneck \cite{gen_bottleneck}. This implies that the genome likely encodes generalized wiring rules rather than specifying individual connections. 

%Another approach for introducing biologically inspired constraints involves embedding neurons of \acp{ANN} in three-dimensional Euclidean space, mimicking the brain's own embedding in 3D space.
Inspired by the brain's embedding in space and its associated spatial constraints, previous work has explored embedding neurons of \acp{ANN} in three-dimensional Euclidean space. Studies have shown that biologically-informed spatial constraints on connection weights lead to connectivity patterns similar to those observed in the brain \cite{Consequence_Bias_Short_Connect, Spatial_Embed_Imposes_Constraints, blauch_connectivity-constrained_2022, seRNN}. However, in these previous approaches, neuron positions remained fixed throughout training and influenced computation only through regularization terms. This raises the question: How would allowing neuron positions to be dynamically optimized during training impact both network structure and function?

To address this question, we propose a parameter-efficient neural architecture in which artificial neurons are embedded within Euclidean space. Their positions are optimized during training, and synaptic weights between connected neurons are computed as the inverse of their distance. By replacing explicit weight matrices with a wiring rule that specifies the connections based on neuron distance, we achieve a more parameter-efficient model representation. In this work, we evaluate this approach for \acp{MLP} and \acp{SNN}.

%% file: sections/methods.tex
We introduce a parameter-efficient neural architecture where each neuron is assigned a position in three-dimensional Euclidean space. During training, these positions are optimized and the connection weights between neuron pairs are computed based on their spatial distances. This approach differs from traditional \acp{ANN}, which learn by optimizing individual connection weights. Assigning each neuron $i$ a position $p_i$ in Euclidean space, the weight of the connection between neurons $i$ and $j$ is then defined as a differentiable function of their distance: 

\begin{equation}
 w_{ij} = \frac{1} {\|p_i - p_j\|_2}.
\end{equation}

\begin{figure} [t]
    \centering
    \includegraphics[width=0.9\linewidth]{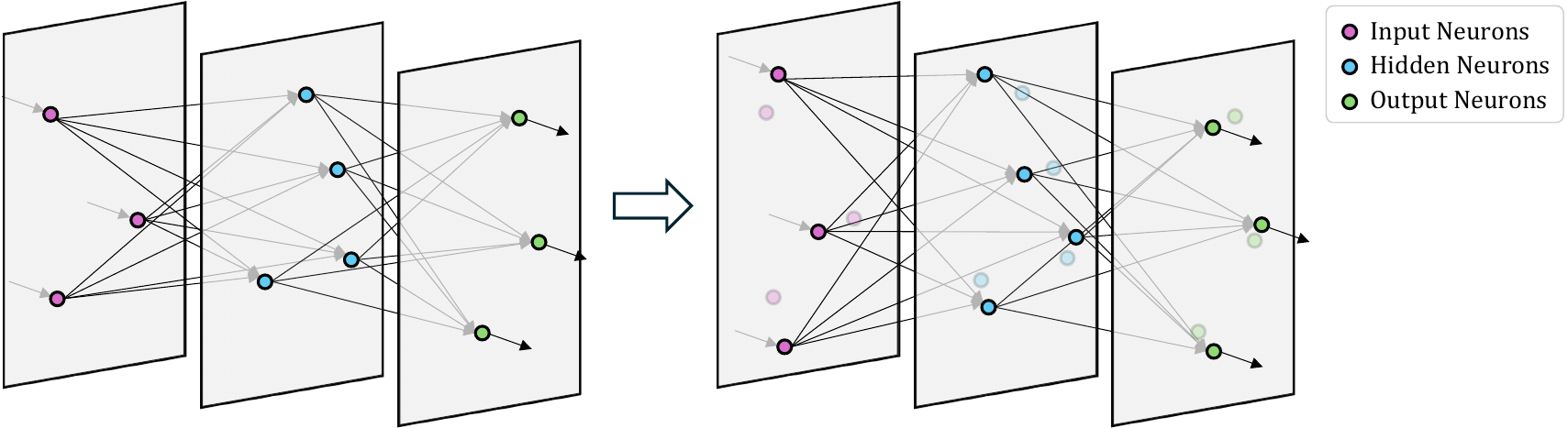}
    \caption{An illustration of a three-layer feedforward network embedded in three-dimensional Euclidean space. Neurons optimize their positions within their respective two-dimensional layers.}
    \label{fig:3D-ANN-structure}
\end{figure}

This formula for the connection weights is inspired by the inverse relation between the conductance $G$, which is analogous to synaptic strength, and the length $L$ of the resistive material in an electrical circuit:
\begin{equation}
 G = \frac{A} {\rho \cdot L}, 
\end{equation}
where $A$ is the cross-sectional area and $\rho$ is the resistivity of the material.

We organize neurons into layers by keeping their z-coordinate fixed to correspond to their layer's index. This ensures a structured feedforward flow, where neurons can optimize their positions within their respective two-dimensional layers, as illustrated in Fig. \ref{fig:3D-ANN-structure}. To enable negative connection weights, we distinguish between excitatory and inhibitory neurons by multiplying their output by a biologically inspired inhibition mask. Each neuron is assigned a continuous inhibition value, which is transformed by a scaled steep sigmoid function to ensure that neurons with positive inhibition values receive an inhibition mask close to 1 (excitatory) and those with negative values a mask close to $-1$ (inhibitory). The inhibition values are optimized during training, allowing the model to flexibly learn the distribution of excitatory and inhibitory neurons. 

Computing the connection weights from neuron positions, rather than explicitly defining the entire weight matrix, significantly reduces the number of parameters in the model while preserving the same number of connections as in traditional architectures. Let $N_L$ be the number of layers, where each layer $l \in \{1, ...,N_L\}$ contains $n_l$ neurons. In a traditional \ac{MLP}, the total number of parameters consists of both the connection weights and the bias terms. It is given by:

\begin{equation}
\sum_{i=1}^{N_L-1}{n_i n_{i+1}} + \sum_{i=2}^{N_L}{n_i}.
\end{equation}

In contrast, a spatially embedded \ac{MLP} optimizes neuron positions instead of the connection weights. Its total number of parameters consists of the neuron coordinates, the bias terms, and the inhibition mask. This can be formulated as:

\begin{equation}
    (d-1) \cdot \sum_{i=1}^{N_L}{n_i} + \sum_{i=2}^{N_L}{n_i} + \sum_{i=1}^{N_L-1}{n_i},
\end{equation}

where $d=3$ is the dimensionality of the metric space in which the neurons are embedded. Each neuron optimizes only $d-1=2$ coordinates, because the z-coordinate remains fixed to correspond to the neuron's layer. With $n$ representing the total number of neurons in the network, the number of parameters in a traditional \ac{MLP} scales as $O(n^2)$, whereas a spatially embedded \ac{MLP} reduces this complexity to $O(n)$. This parameter-efficient framework can be combined with existing compression techniques such as pruning or quantization, thereby complementing and extending prior work.

%% file: sections/results.tex
The proposed concept can be integrated into different neural architectures, such as \acp{MLP} or \acp{RNN}, without requiring fundamental changes to their core design. In this work, we specifically focus on embedding \acp{MLP} and fully connected \acp{SNN} in three-dimensional Euclidean space. For training and evaluation, we use the MNIST \cite{MNIST} dataset. All models in this work have three layers, with 10 input neurons and 784 output neurons. Models were trained with three different random seeds, with cross-entropy loss as the objective function. Training is performed in mini-batches of size 600 using the Adam optimizer \cite{adam_opt}. The SNNs are implemented using Leaky Integrate-and-Fire (LIF) neurons from the SNNTorch library \cite{SNN_Review_Deep_Learning}, with a threshold value of 1 and a beta value of 0.95, which are not modified during training.

%Additional experiments that did not improve performance can be found in Appendices \ref{sup:weighted_embedding} and \ref{sup:input_grid}.

\subsection{Performance Evaluation}
\label{sec:Experiments_Performance}

\begin{figure}
\begin{subfigure}[b]{0.48\textwidth}
    \includegraphics[width=\linewidth]{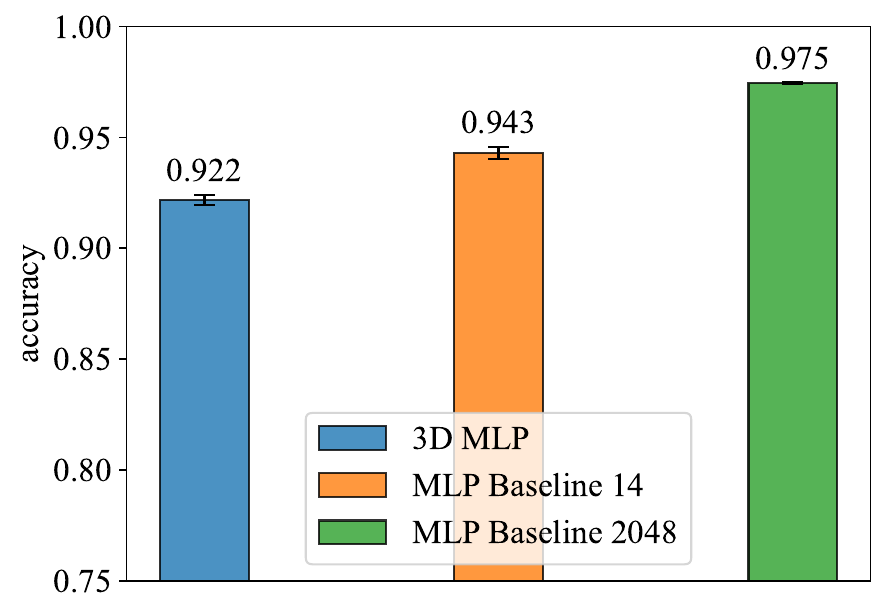}
    \caption{MLP Performance Comparison}
    \label{fig:ANN_Performance_Comp}
  \end{subfigure}
\begin{subfigure}[b]{0.48\textwidth}
    \includegraphics[width=\linewidth]{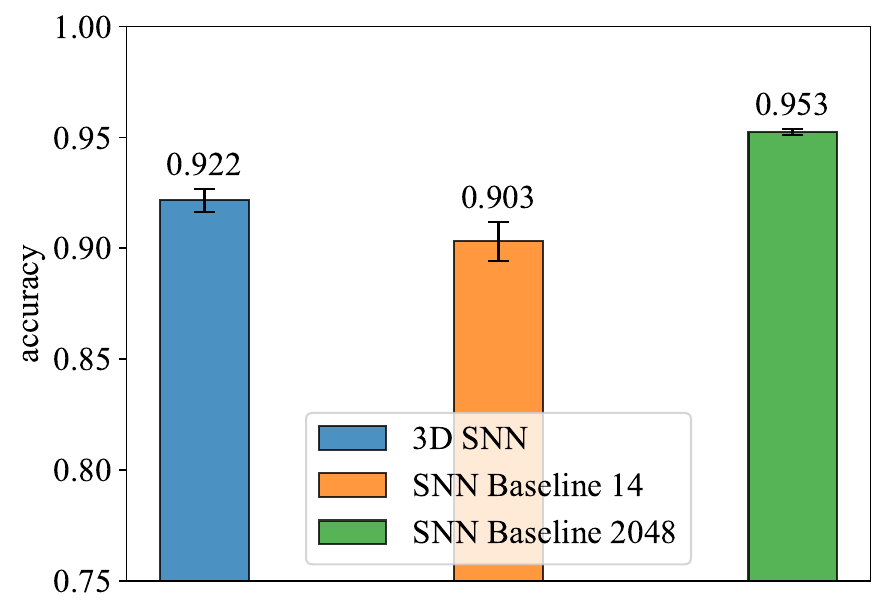}
    \caption{SNN Performance Comparison}
    \label{fig:SNN_Performance_Comp}
  \end{subfigure}
    \caption{Accuracy comparison of the (\subref{fig:ANN_Performance_Comp}) 3D MLP and (\subref{fig:SNN_Performance_Comp}) 3D SNN (with 2,048 hidden neurons) with the baseline models (with 14 and 2,048 hidden neurons) on the MNIST \cite{MNIST} dataset. Models were trained using three different random seeds, with error bars representing the standard deviation across runs.}
    \label{fig:Performance_Comp}
\end{figure}

We implement an \ac{MLP} embedded in a three-dimensional Euclidean space with fixed layer distances set to one across all layers. It has a single hidden layer with 2,048 neurons, resulting in a total of 10,574 learnable parameters. After being trained for 300 epochs with a learning rate of 0.005, the model achieves a test accuracy of 0.9018 $\pm$ 0.0042. Next, we relax the constraint that all layers must share the same fixed layer distance and implement a \dddANNLearnedLayerDist with independently learned layer distances, referred to as \dddANNLearnedLayerDist. This model achieves an improved test accuracy of 0.9217 $\pm$ 0.0024 with only a minor increase in the number of parameters (10,576).
We compare the performance of the spatially embedded MLPs to two MLP baselines, which have the same architecture but vary in the number of hidden neurons. They are designed to match either the number of parameters or number of hidden neurons to more accurately assess the affect of the distance dependent wiring rule. \ANNBaselineX 2048 has 2,048 hidden neurons to match the number of neurons in the \dddANNLearnedLayerDist, but contains significantly more parameters. \ANNBaselineX 14 is a more compact MLP with only 14 hidden neurons, designed to have a parameter count comparable to that of the \dddANNLearnedLayerDist. After training for 300 epochs with learning rate 0.001, the two baselines outperform the spatially embedded MLPs (see Table \ref{table:ANN_Performance_Comp} and Fig. \ref{fig:ANN_Performance_Comp}).

\begin{table} [t]
\centering
\caption{Accuracy comparison of the 3D MLP (with 2,048 hidden neurons) and the baseline MLPs (with 14 and 2,048 hidden neurons) on the MNIST \cite{MNIST} dataset. Models were trained using three different random seeds and accuracy is reported as the mean ± standard deviation.}
\begin{tabular}{ l | c | c | c}
 Model & Test Accuracy & Number of Parameters & Learning Rate\\
  \hline\hline
 \dddANNLearnedLayerDist & 0.9217 $\pm$ 0.0024 & 10,576 & 0.005\\ 
 \hline
 \ANNBaselineX 14 & 0.9429 $\pm$ 0.0028 & 11,140 & 0.001 \\
 \ANNBaselineX 2048 & 0.9745 $\pm$ 0.0003 & 1,628,170 & 0.001
\end{tabular}
\label{table:ANN_Performance_Comp}
\end{table}

We further evaluate the spatial embedding framework for fully connected SNNs. Analogous to the evaluation of the 3D MLPs, we train an SNN embedded in 3D space with 2,048 hidden neurons with fixed layer distances. After training for 200 epochs with learning rate 0.005, the model achieves a test accuracy of 0.9187 $\pm$ 0.0027. 
Allowing the layer distances to be independently learnable for each layer slightly improves the performance of the 3D SNN to 0.9216 ± 0.0052. 
When compared to conventional fully connected SNNs trained for 200 epochs with learning rate 0.001, \SNNBaselineX 2048 with 2,048 hidden neurons outperforms the 3D SNNs, while having the same number of weights but significantly more parameters. In contrast, the 3D SNNs outperform the \SNNBaselineX 14, which has only 14 hidden neurons and a number of parameters comparable to that of the 3D SNNs (see Table \ref{table:SNN_Performance_Comp} and Fig. \ref{fig:SNN_Performance_Comp}).

\begin{table}
\centering
\caption{Accuracy comparison of the 3D SNN (with 2,048 hidden neurons) and the baseline SNNs (with 14 and 2,048 hidden neurons) on the MNIST \cite{MNIST} dataset. Models were trained using three different random seeds, and accuracy is reported as the mean ± standard deviation.}
\begin{tabular}{ l | c | c | c}
 Model & Test Accuracy & Number of Parameters & Learning Rate \\
  \hline\hline
 \dddSNNLearnedLayerDist & 0.9216 $\pm$ 0.0052 & 10,576 & 0.005 \\ 
 \hline
 \SNNBaselineX 14 & 0.9031 $\pm$ 0.0088 & 11,140 & 0.001 \\
 \SNNBaselineX 2048 & 0.9525 $\pm$ 0.0013 & 1,628,170 & 0.001
\end{tabular}
\label{table:SNN_Performance_Comp}
\end{table}

\subsection{Magnitude-Based Weight Pruning}
\label{sec:Experiment_Pruning}

Mimicking the brain’s tendency to minimize wiring length \cite{Minimum_wiring_length}, we prune the longest connections within our spatially embedded models. Since the connection weights are the inverse of the distance between neurons, this is equivalent to removing the connections with the smallest weights, effectively implementing magnitude-based weight pruning. 

\begin{figure} [t]
  \begin{subfigure}[b]{0.48\textwidth}
    \includegraphics[width=\textwidth]{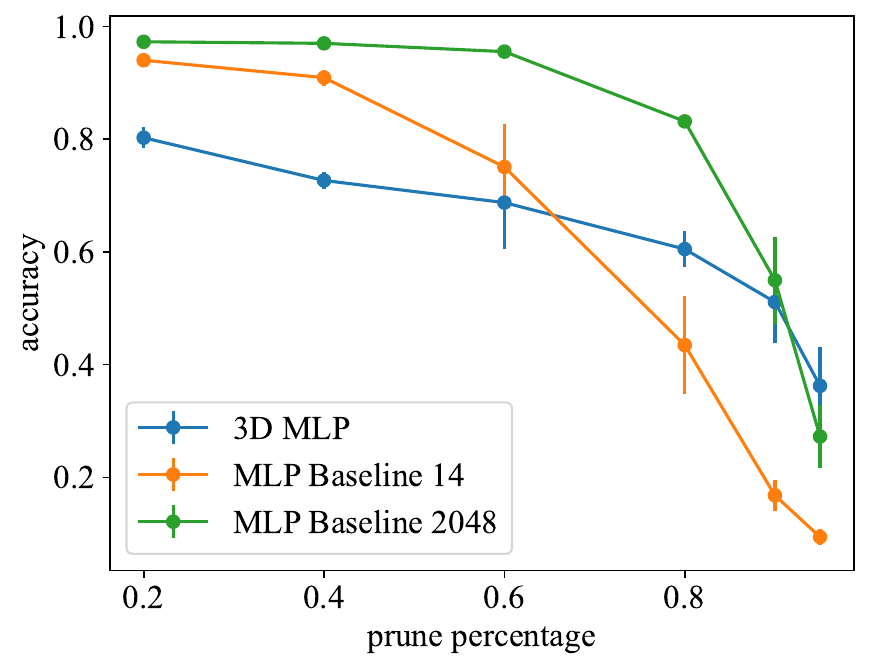}
    \caption{Models are pruned after training.}
    \label{fig:prune_acc_tradeOff_ANN_afterTrain}
  \end{subfigure}
  \hfill
  \begin{subfigure}[b]{0.48\textwidth}
    \includegraphics[width=\textwidth]{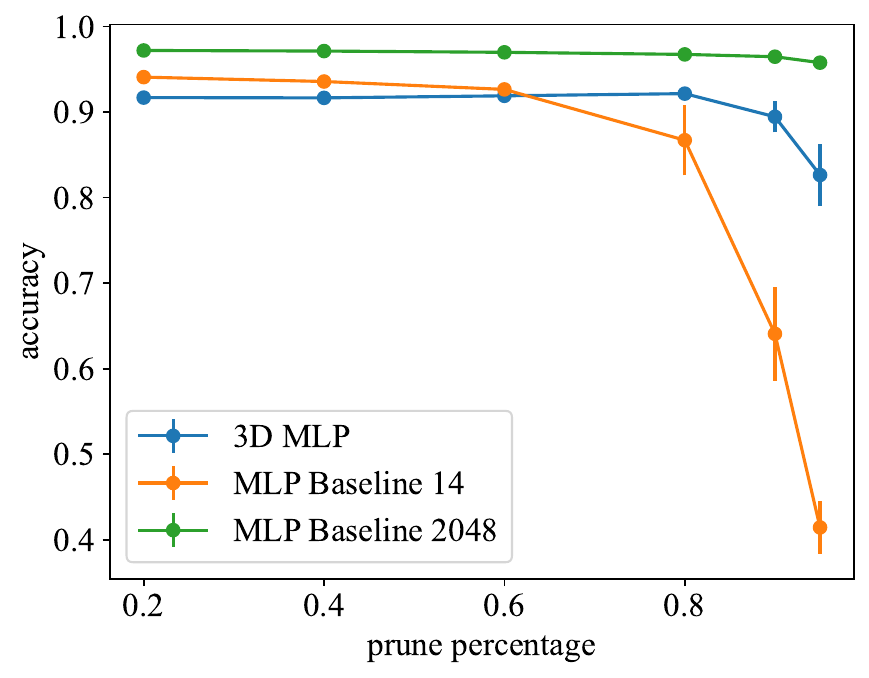}
    \caption{Models are pruned during training.}
    \label{fig:prune_acc_tradeOff_ANN_duringTrain}
  \end{subfigure}
  \caption{Accuracy comparison of the 3D MLP (with 2,048 hidden neurons) and the baseline MLPs (with 14 and 2,048 hidden neurons) on the MNIST \cite{MNIST} dataset for different pruning percentages. The models are pruned (\subref{fig:prune_acc_tradeOff_ANN_afterTrain}) after training and (\subref{fig:prune_acc_tradeOff_ANN_duringTrain}) during training by setting the smallest percentage of weights to zero. Models were trained using three different random seeds, with error bars representing the standard deviation across runs.}
\end{figure}

We first apply post-training pruning, where weights are pruned only after the model has been fully trained. We evaluate this approach for the models that were introduced in the previous Section \ref{sec:Experiments_Performance}, namely the \dddANNLearnedLayerDist with 10,576 parameters and 1,626,112 weights, \ANNBaselineX 14 with 11,140 parameters and 11,116 weights, and \ANNBaselineX 2048 with 1,628,170 parameters and 1,626,112 weights. For the baselines, removing connections is equivalent to removing parameters. The results of post-training pruning are visually presented in Fig. \ref{fig:prune_acc_tradeOff_ANN_afterTrain}. Initially, \ANNBaselineX 14 outperforms the \dddANNLearnedLayerDist, but for pruning percentages of 80\% or higher, the \dddANNLearnedLayerDist achieves better performance. However, despite both models having the same number of parameters, \ANNBaselineX 14 contains significantly fewer hidden neurons and connection weights. For a fair comparison, \ANNBaselineX 2048 has the same number of connection weights as the \dddANNLearnedLayerDist. It generally outperforms the spatially embedded MLP, but at an extreme pruning percentage of 95\%, the \dddANNLearnedLayerDist outperforms even the \ANNBaselineX 2048, which has significantly more parameters.

In the second approach, pruning is integrated into the training process. In each forward pass, the smallest connection weights are identified and removed. We apply this strategy to \dddANNLearnedLayerDist, \ANNBaselineX 14, and \ANNBaselineX 2048. The results of pruning during training are illustrated in Fig. \ref{fig:prune_acc_tradeOff_ANN_duringTrain}. 
As observed in post-training pruning, \ANNBaselineX 14 initially outperforms the \dddANNLearnedLayerDist. 
However, for pruning percentages above 60\%, the \dddANNLearnedLayerDist surpasses the \ANNBaselineX 14, which has fewer connections and neurons but the same number of parameters. When pruning during training, \ANNBaselineX 2048 consistently outperforms \dddANNLearnedLayerDist while also having significantly more parameters.

\subsection{Optimizing Z-Coordinates}

\begin{figure}
  \begin{subfigure}[h]{0.48\textwidth}
    \vspace{-.5cm}
    \includegraphics[width=\linewidth]{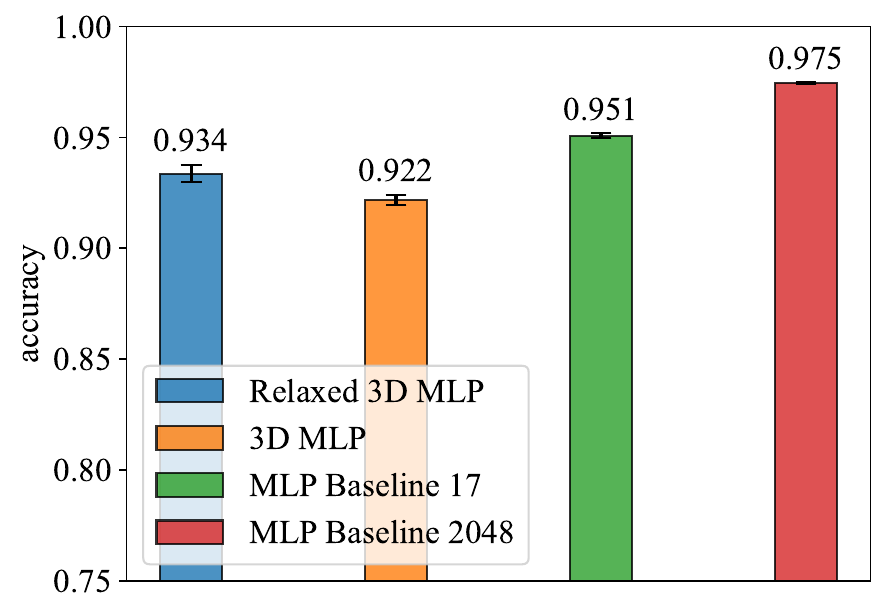}
    \caption{}
    \label{fig:FreeZPerformance}
  \end{subfigure}
  \hfill
  \begin{subfigure}[h]{0.48\textwidth}
    \includegraphics[width=\textwidth]{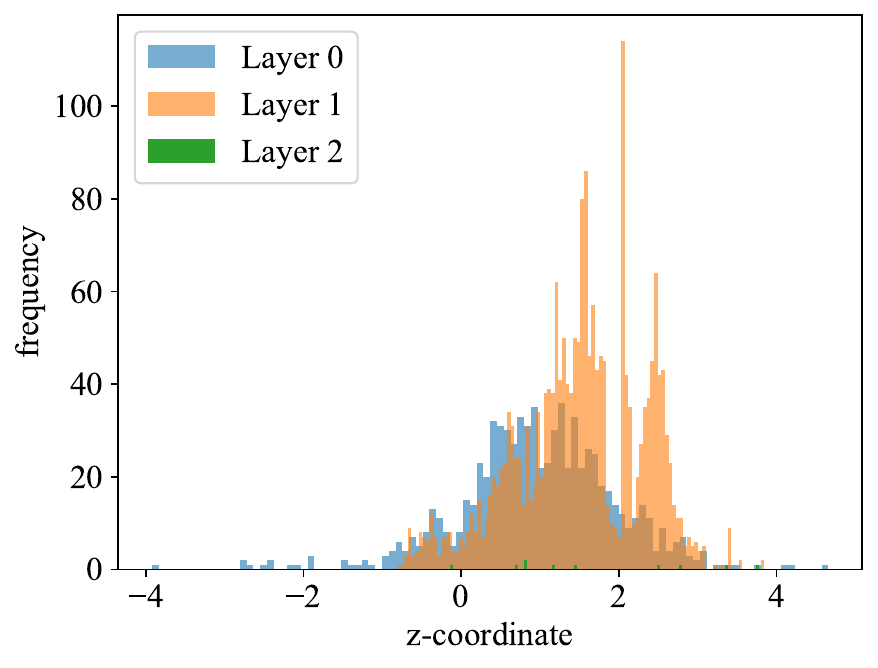}
    \caption{}
  \label{fig:Experiment:z_distr}
  \end{subfigure}
  \caption{(\subref{fig:FreeZPerformance})Accuracy comparison of the \dddANNfreeZ{}s with learned z-coordinates, the \dddANNLearnedLayerDist, and the baseline MLPs (with 17 and 2,048 hidden neurons) on the MNIST \cite{MNIST} dataset. Models were trained using three different random seeds, with error bars representing the standard deviation across runs.
  (\subref{fig:Experiment:z_distr}) The distribution of the z-coordinates in the \dddANNfreeZ{} after training. }
\end{figure}

While structuring neurons into spatial layers provides a simple and organized framework, it limits the model's expressiveness by constraining neuron placement. 
Therefore, we relax the spatial layer hierarchy and allow neurons to optimize their z-coordinate during training while preserving the layer-based connectivity for computational simplicity. This relaxation adds an additional learnable parameter, namely the z-coordinate, for each neuron. The z-coordinates are initialized layerwise in the range [$l$, $l + 1$) for layer $l$. We train a relaxed spatially embedded MLP with 2,048 neurons, referred to as \dddANNfreeZ, for 300 epochs with learning rate 0.005. The model achieves a test accuracy of 0.9337 $\pm$ 0.0040, representing an improvement over \dddANNLearnedLayerDist. However, it is still outperformed by \ANNBaselineX 17, which has approximately the same number of parameters but only 17 hidden neurons (see Table \ref{table:FreeZPerformance} and Fig. \ref{fig:FreeZPerformance}). Figure \ref{fig:Experiment:z_distr} shows the distribution of the learned z-coordinates for the \dddANNfreeZ{}. Notably, the z-distributions of all layers exhibit significant overlap.

\begin{table} [t]
\centering
\caption{Accuracy comparison of the \dddANNfreeZ{}s with learned z-coordinates, the \dddANNLearnedLayerDist, and the baseline MLPs (with 17 and 2,048 hidden neurons) on the MNIST \cite{MNIST} dataset. Models were trained using three different random seeds, and accuracy is reported as the mean ± standard deviation.}
\begin{tabular}{ l | c | c | c}
 Model & Test Accuracy & Number of Parameters & Learning Rate \\
  \hline\hline
 \dddANNfreeZ & 0.9337 $\pm$ 0.0040 & 13,416 & 0.005 \\
 \hline
 \dddANNLearnedLayerDist & 0.9217 $\pm$ 0.0024 & 10,576 & 0.005 \\
 \ANNBaselineX 17 & 0.9507 $\pm$ 0.0011 & 13,525 & 0.001 \\
 \ANNBaselineX 2048 & 0.9745 $\pm$ 0.0003 & 1,628,170  & 0.001
\end{tabular}
\label{table:FreeZPerformance}
\end{table}

\subsection{Higher-Dimensional Embedding Spaces} 
%Increasing the dimensionality of the embedding space provides additional degrees of freedom, allowing neurons to form more complex geometric relationships. 

\begin{figure}
    \centering
    \includegraphics[width=0.8\linewidth]{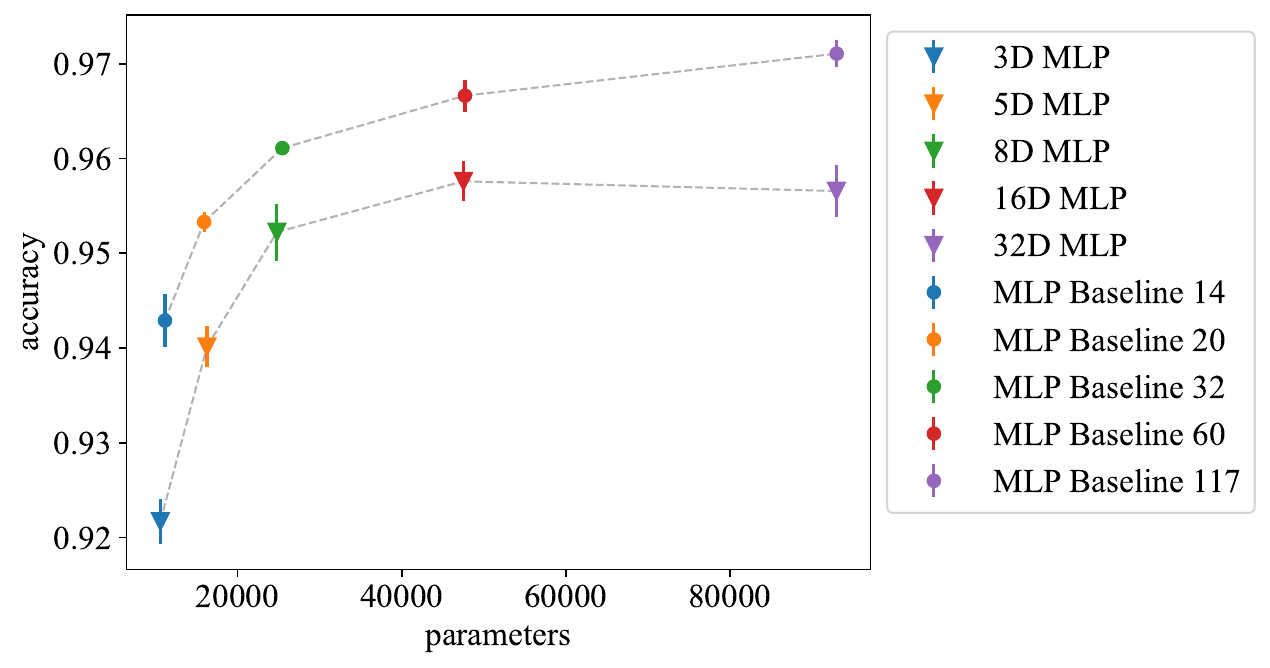}
    \caption{Accuracy comparison of MLPs with 2,048 hidden neurons embedded in 3-, 5-, 8-, 16-, and 32-dimensional space, and the baseline MLPs with 14, 20, 32, 60, and 117 hidden neurons on the MNIST \cite{MNIST} dataset. Models were trained using three different random seeds, with error bars representing the standard deviation across runs.}
\label{fig:Experiment:MoreDims_Performance_Comp}
\end{figure}

From a graph-theoretic perspective, increasing the dimensionality of the embedding space provides additional degrees of freedom. Although this departs from the motivation to mimic the brain's embedding in 3D space, higher-dimensional spaces allow neurons to form more complex geometric relationships. We embedded models in 5-, 8-, 16-, and 32-dimensional Euclidean spaces, each with 2,048 hidden neurons. Increasing the dimensionality also increases the number of parameters, so we compared the models' performance to MLP baselines with similar parameter counts with 20, 32, 60, and 117 hidden neurons, respectively. Notably, the models embedded in higher-dimensional spaces still have fewer parameters than the baseline with the same number of neurons. Although increasing the number of dimensions of the embedding space improves performance, the spatially embedded models are still outperformed by their conventional MLP counterparts (see Fig. \ref{fig:Experiment:MoreDims_Performance_Comp}).

% \begin{table}
% \centering
% \resizebox{\textwidth}{!}{%
% \begin{tabular}{ c | c | c || c | c | c }
%  Model & Test Accuracy & Params & Model & Test Accuracy & Params\\
%   \hline\hline
%   \dddANNLearnedLayerDist & 0.9217 $\pm$ 0.0024 & 10,576 &     
%   \ANNBaselineX 14 & 0.9429 $\pm$ 0.0028 &  11,140  \\
%   5D ANN & 0.9401 $\pm$ 0.0022 &  16,260 & \ANNBaselineX 20 & 0.9533 $\pm$ 0.0011 &   15,910\\  
%   8D ANN & 0.9522 $\pm$ 0.0030 &  24,786 & \ANNBaselineX 32 & 0.9611 $\pm$ 0.0007 & 25,450 \\  
%   16D ANN & 0.9576 $\pm$ 0.0021 & 47,522 & \ANNBaselineX 60 & 0.9666 $\pm$ 0.0017 &  47,710 \\  
%   32D ANN & 0.9566 $\pm$ 0.0027 &  92,994 & \ANNBaselineX 117 & 0.9711 $\pm$ 0.0014 &  93,025\\  
%  \hline
%   %\ANNBaselineX 20 & 0.9533 $\pm$ 0.0011 &   15,910 \\
%   %\ANNBaselineX 32 & 0.9611 $\pm$ 0.0007 & 25,450   \\
%   %\ANNBaselineX 60 & 0.9666 $\pm$ 0.0017 &  47,710  \\
%   %\ANNBaselineX 117 & 0.9711 $\pm$ 0.0014 &  93,025  \\
%  %  \dddANNLearnedLayerDist & 0.9217 $\pm$ 0.0024 & 10,576 \\     
%  %  \ANNBaselineX 14 & 0.9429 $\pm$ 0.0028 &  11,140  \\
%  % \ANNBaselineX 2048 & 0.9745 $\pm$ 0.0003 & 1,628,170 
% \end{tabular}
% }
% \caption{Accuracy comparison of MLPs embedded in 3-, 5-, 8-, 16-, and 32-dimensional space, and the baseline MLPs with 14, 20, 32, 60, 117 and 2048 hidden neurons on the MNIST \cite{MNIST} dataset. Models were trained using three different random seeds, and accuracy is reported as the mean ± standard deviation.}
% \label{table:Experiment:MoreDims_Performance_Comp}
% \end{table}

%% file: sections/discussion.tex
Spatially embedded models demonstrated competitive performance. Notably, the spatially embedded SNN outperformed the spiking baseline model with a comparable parameter count, indicating a more efficient parameter utilization. This particularly strong performance of the spatially embedded SNN likely stems from its ability to leverage temporal dynamics for information encoding. This allows the model to compensate for the reduced weight flexibility through precise spike timing. In contrast, MLPs rely primarily on weight adjustments and may face challenges with restricted connectivity.

% === free Z ======
Omitting the spatial layer structure improved performance, likely because it allowed the model to strengthen connections between specific neurons by moving them closer together. The significant overlap of z-coordinate distributions across layers confirms that the models leveraged this additional flexibility.
% ======== More Dimensions ============
Increasing the dimensionality of the embedding space improved performance, while the embedded models maintained a lower parameter count compared to conventional models with the same number of neurons.
% Models embedded in higher-dimensional spaces Increasing the dimensionality of the embedding space further improved performance by providing additional degrees of freedom for neuron placement, while maintaining better parameter-efficiency than baseline models with the same number of neurons. 
The performance plateaued beyond 16 dimensions, indicating that only a limited number of dimensions are required to express the weight configurations necessary for complex computations.

% ======= Pruning ==============
Spatially embedded models further demonstrated robustness to pruning. At sparsity levels above 60\%, they outperformed baselines with similar parameter counts under the same conditions. When pruned after training with an extreme pruning percentage of 95\%, the 3D MLP even outperformed the baseline with the same number of connection weights but significantly more parameters. Notably, when pruning 80\% of the connections during training, the performance of the 3D MLP slightly improved again, nearly matching the accuracy of the unpruned model. These findings can be explained by the distance-dependent weight computation. Since distances cannot be infinite, the spatial embedding inherently prevents connections from having zero weights. As a result, each neuron receives input from all neurons in the preceding layer, even if some of these inputs are noisy or irrelevant. Pruning the smallest weights effectively enables zero-weighted connections, eliminating those that would otherwise introduce noise. Additionally, pruning reduces the dependencies between connection weights imposed by spatial embedding. Since modifying the position of a single neuron affects all its associated connection weights, removing less critical connections allows neurons to adjust positions more freely with fewer unintended side effects.

\begin{figure} [t]
        \begin{subfigure}{0.326\textwidth}
            \includegraphics[width=\textwidth]{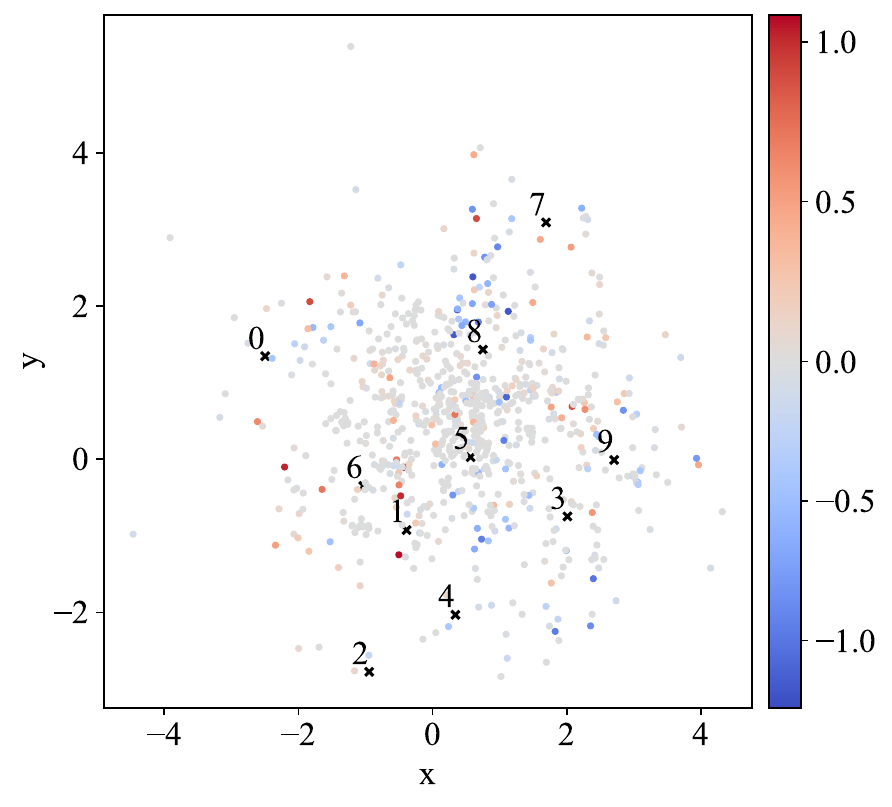}
            \caption{Input Neurons \\}
        
        \end{subfigure}
        \begin{subfigure}{0.32\textwidth}
            \includegraphics[width=\textwidth]{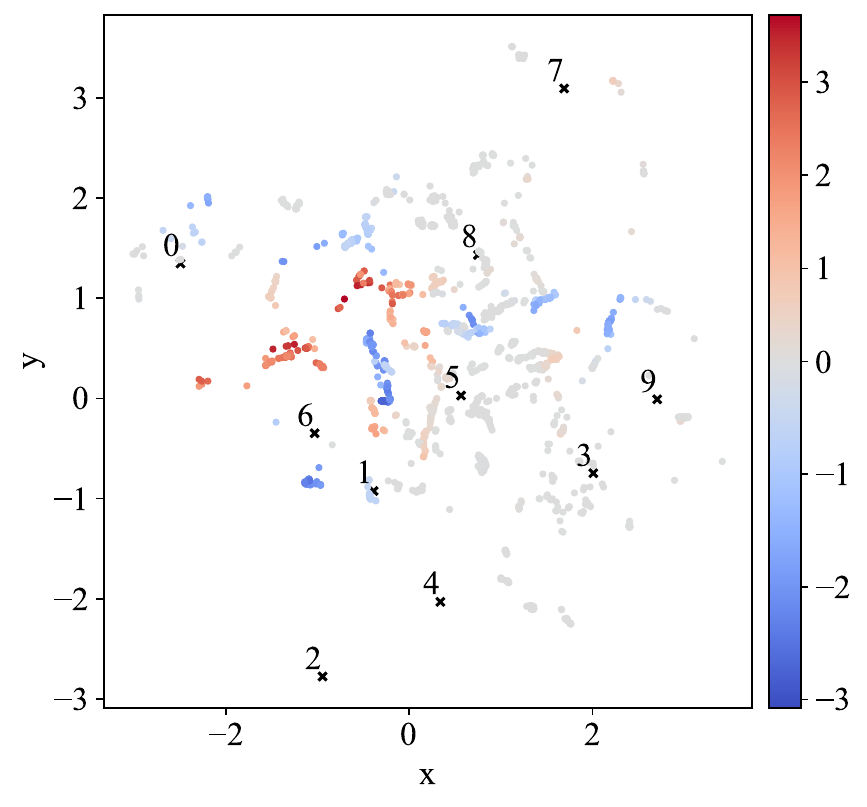}
            \caption{Hidden Neurons \\}
            
        \end{subfigure}
        \begin{subfigure}{0.32\textwidth}
            \includegraphics[width=\textwidth]{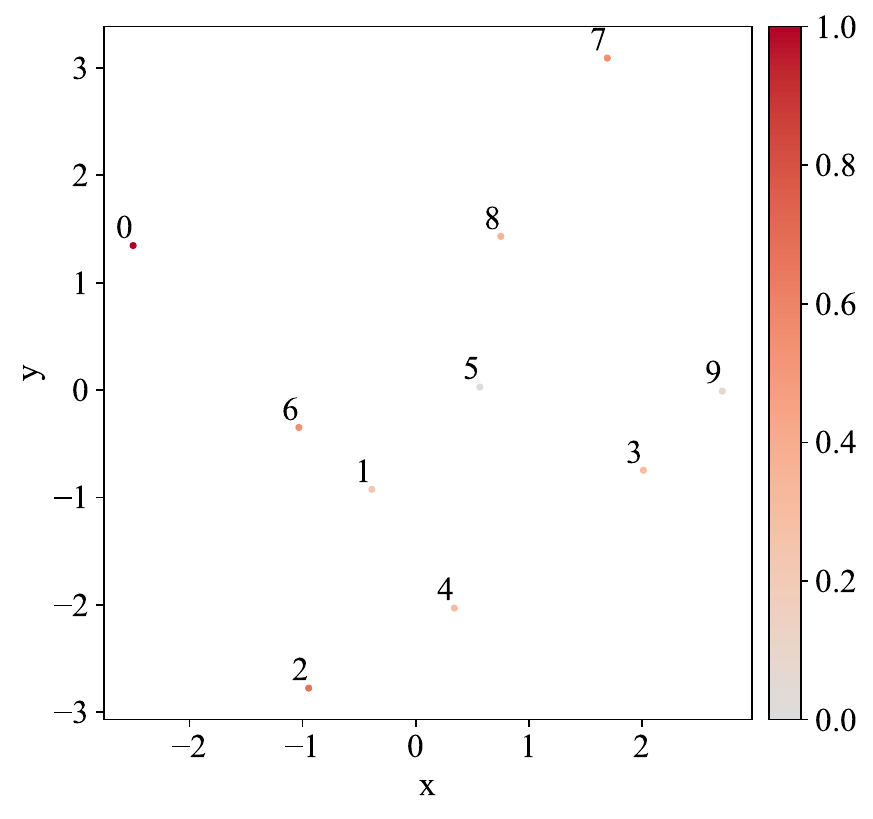}
            \caption{Output Neurons \\}
            
        \end{subfigure}
        
        \caption{The final neuron positions of a spatially embedded MLP with three layers. The color indicates the strength of the neurons' activation in response to the average input image for digit 0. Blue indicates inhibitory activation while red indicates excitatory activation. The positions of the output neurons are also included in the visualizations of the input and hidden layer as reference points.
}
       \label{fig:discussion:NeuronActDigit0}
\end{figure}

Making neuron positions learnable and deriving connection weights from their distances establishes a direct link between geometry and connectivity. As a result, spatial constraints are naturally integrated into the architecture itself. Similarly to neural circuits in the brain, which evolve under pressure to minimize wiring length \cite{Minimum_wiring_length}, the proposed model inherently favors spatially compact connectivity patterns by assigning stronger weights to connections between nearby neurons.
% ========= Neuron Visualization ====
Moreover, by structuring neurons in 2D layers, the spatial embedding framework offers an intuitive way of visualizing the spatial arrangement of neurons and their activations, as illustrated in Fig. \ref{fig:discussion:NeuronActDigit0}. %and further detailed in Appendix \ref{sec:Experiment:NeuronActivationVisual}. 
These visualizations help explain why the model makes certain decisions and identify neurons that are essential for particular features. Beyond interpretability, this framework is also useful for debugging. Outlier neurons positioned far from input or output clusters may indicate underutilized components, enabling targeted architectural adjustments. 

Although spatially embedded MLPs demonstrated competitive performance, they were consistently outperformed by conventional MLPs, even when compared to baseline models with a similar number of parameters but significantly fewer neurons. This performance gap can be attributed to the dependencies between connection weights. Since the connection weights are determined by distance, modifying the position of a single neuron affects all connection weights linked to that neuron. This interdependence restricts the range of possible weight configurations, as adapting individual connection weights is not possible. As a consequence, the spatial embedding limits the model's expressiveness and likely prevents spatially embedded MLPs from outperforming traditional models when evaluated on the MNIST dataset. However, the spatial embedding can be a useful constraint in other specific tasks and future research is needed to investigate the effectiveness of spatially embedded models on more diverse datasets.

%% file: sections/conclusion.tex
Motivated by the need for resource-efficient models with a low memory footprint and to bridge the gap between artificial and biological neural systems, we proposed a neural architecture in which neurons are embedded in three-dimensional Euclidean space, mimicking the spatial embedding of the brain. By optimizing neuron positions rather than individual connection weights, we reduced the model's parameter complexity from $O(n^2)$ to $O(n)$, where $n$ is the number of neurons in the model. 
% ====== Conclusion based on Analysis ==========
Spatially embedded models achieved competitive performance compared to baseline MLPs and even outperformed the baseline SNN with the same number of parameters. Increasing the embedding dimensionality or relaxing the spatial layer hierarchy further improved performance. Moreover, the spatially embedded architecture demonstrated robustness to pruning and maintained accuracy even when 80\% of the weights were pruned during training -- a valuable property for deployment on resource-constrained devices. Beyond efficiency, visualizing neuron positions and their activations offers deeper insights into the inner workings of spatially embedded neural networks. 

% ======== Future work ============
The current experimental validation is limited to shallow networks trained on the MNIST dataset. To assess the scalability of the spatial embedding framework and to evaluate its performance in more challenging scenarios, future work should investigate its application to deeper architectures and more complex, diverse datasets. Moreover, a promising direction for future research is to explore more flexible spatial embeddings that better emulate biological neural systems. For example, relaxing rigid layer-wise connectivity structure and allowing the network to learn its connectivity could provide greater flexibility. Introducing sparse long-range connections, similar to the brain’s heavy-tailed degree distributions, could further increase model expressiveness while maintaining efficiency through structural constraints. 
%Since the experiments were conducted on shallow networks for MNIST classification, extending the spatial embedding framework to deeper architectures and more diverse datasets will be essential to assess its scalability. 
Furthermore, since the application to SNNs proved to be successful, the spatial embedding approach may also be well-suited for other neural architectures, such as Graph Neural Networks (GNNs), where input data often naturally exhibit spatial or topological structure.
% === Summary ====
In summary, this work established spatially embedded neural networks as a novel framework for designing compact and biologically inspired AI systems.